\documentclass[a4paper, 10pt, conference]{IEEEtran}
\IEEEoverridecommandlockouts
\usepackage{cite}
\usepackage{amsmath,amssymb,amsfonts}
\usepackage{algorithmic}
\usepackage{graphicx}
\usepackage{textcomp}
\usepackage{xcolor}
\usepackage{tabularx}
\usepackage{bm}
\usepackage{booktabs}
\usepackage{caption}
\usepackage{subcaption}

\def\BibTeX{{\rm B\kern-.05em{\sc i\kern-.025em b}\kern-.08em
    T\kern-.1667em\lower.7ex\hbox{E}\kern-.125emX}}

\begin{document}

\title{\LARGE \bf EEG2Vec: Learning Affective EEG Representations via\\Variational Autoencoders}

\author{David~Bethge$^{1,2,*}$, Philipp~Hallgarten$^{1,3}$, Tobias~Grosse-Puppendahl$^{1}$, Mohamed~Kari$^{1}$, \\Lewis~L.~Chuang$^{4}$, Ozan~\"{O}zdenizci$^{5,6}$, Albrecht~Schmidt$^{2}$
\thanks{$^{1}$Dr. Ing. h.c. F. Porsche AG, Stuttgart, Germany}%
\thanks{$^{2}$Ludwig Maximilian University, Munich, Germany}%
\thanks{$^{3}$University of T\"{u}bingen, T\"{u}bingen, Germany}%
\thanks{$^{4}$Chemnitz University of Technology, Chemnitz, Germany}
\thanks{$^{5}$Institute of Theoretical Computer Science, TU Graz, Austria}%
\thanks{$^{6}$TU Graz-SAL DES Lab, Silicon Austria Labs, Graz, Austria}%
\thanks{$^{*}$Corresponding author: {\tt\small david.bethge@ifi.lmu.de}}%
}

\maketitle

\begin{abstract}
There is a growing need for sparse representational formats of human affective states that can be utilized in scenarios with limited computational memory resources. We explore whether representing neural data, in response to emotional stimuli, in a latent vector space can serve to both predict emotional states as well as generate synthetic EEG data that are participant- and/or emotion-specific. We propose a conditional variational autoencoder based framework, EEG2Vec, to learn generative-discriminative representations from EEG data. Experimental results on affective EEG recording datasets demonstrate that our model is suitable for unsupervised EEG modeling, classification of three distinct emotion categories (positive, neutral, negative) based on the latent representation achieves a robust performance of 68.49$\%$, and generated synthetic EEG sequences resemble real EEG data inputs to particularly reconstruct low-frequency signal components. Our work advances areas where affective EEG representations can be useful in e.g., generating artificial (labeled) training data or alleviating manual feature extraction, and provide efficiency for memory constrained edge computing applications.
\end{abstract}


\section{Introduction}
\label{sec:introduction}

The emphasis on human-centric computing in recent years has accelerated efforts in affective computing to develop effective computer-aided approaches for recognizing, interpreting, processing, and simulating a person's emotions. 
Recent notable successes include affect-adaptive robot-child feedback in education \cite{tielman2014adaptive}, mobile real-time facial emotion annotation systems \cite{zhang2020rcea}, as well as detecting and influencing drivers' emotions \cite{hassib2019detecting}. 
In particular, the growing availability of high-resolution wearable physiological measurement systems has, in combination with powerful machine learning methods, increased recognition performance of affective user states for various applications in-the-wild \cite{hu2019ten}. 

In this paper, we focus on electroencephalographic (EEG) measurements that are elicited in response to affective emotional stimuli. 
Primarily we seek to determine whether a user-specific affective representation from raw EEG can be learned in an end-to-end \textit{representation learning} framework. 
In such a setting, representations of affective states are learned from input data---typically by transforming it or extracting informative features from it (the useful vantage point of the data's key qualities)---towards the objective of performing particular tasks like prediction of affective states from noisy EEG data. 
Traditional discriminative machine learning approaches have the sole objective of classifying distinct affective categories. 
Differently, our focus on representation learning aims at estimating a powerful abstraction of affect-relevant multi-channel sensor input. 
This approach encodes the signal generative components from the training data distribution in a learned latent space.
Once such models are learned, semi-supervised learning schemes can be adopted by e.g., applying the representation learning encoder on unlabeled data for class-conditional data synthesis, which can be combined with labeled data representations for a larger training base to, for instance, predict emotions.

Our work is inspired by recent advances in word representations, also denoted as \textit{embeddings} \cite{fei2020latent,ruan2021emotion}. 
A prominent success story is \textit{word2vec} \cite{mikolov2013distributed} in natural language processing, which uses a neural network model to learn word representations from a large text corpus. 
Once trained, such representation models can detect synonymous words or accurately suggest additional words for a partial sentence. 
This has given rise to numerous natural language applications that were previously unimaginable (e.g., predicting the right next words in chats, sentiment analysis of messages, machine translation, click session advertisement-recommendation, automatic topic clustering). 
State-of-the-art natural language processing algorithms can even learn cross-lingual concepts~\cite{williams2021exploring}, generate complete texts~\cite{devlin2018bert,brown2020language} or infer emotion-related text sentiments~\cite{yang2017improved}. 
When trained with enough data, word \textit{embeddings} tend to capture word concepts and meanings, and are even able to perform analogies, e.g., the vector for ``Paris" minus the vector for ``France" plus the vector for ``Italy" is very close to the vector for ``Rome". 
Thereby, word representations can bridge the human understanding of language to that of a machine. 

We seek to contribute towards ubiquitous wearable physiological systems and allow for computing systems that are responsive to user affective states in real-world scenarios. 
Here, we focus on high-dimensional EEG data and share how a highly informative and compressed representation could be derived from it to support this vision. 
Once learned, such representations can be useful for downstream tasks, such as predicting emotions to better understand the affective states in the brain through representations in a lower dimensional space, or simulating synthetic EEG data. 

To date, affective brain-computer interface (BCI) methodologies are often hindered by the lack of large labeled datasets, low signal-to-noise (SNR) ratio in real-time EEG data acquisition, or non-reproducible handcrafted feature extraction \cite{saha2019progress}. 
Our work investigates variational representation learning for affective EEG data~\cite{zheng2015investigating}, that is able to mitigate these issues by learning a suitable and easy-to-use emotion representation trained for data augmentation and emotion recognition. 
Furthermore, many in-the-wild affective applications share the common obstacle of efficiently processing raw EEG data \cite{saha2019progress, lupu2019brain}, which can be reduced by processing an information condensed latent space vector, which ultimately can be also streamed to the cloud with bandwidth restrictions. 

We term our proposed model \textit{EEG2Vec}, which exploits a conditional variational autoencoder (cVAE) \cite{kingma2013auto} structure by encoding raw multi-channel EEG signals into a shared latent vector space while a simple feed-forward emotion classification neural network is simultaneously harnessing these latent representations as input. 
Subsequently, the resulting latent representations can be used to (1) predict the affective state of user from their current EEG recordings within a discriminative framework, as well as (2) to generate emotion- and subject-specific synthetic multi-channel EEG signals.

\section{Related Work}
\label{sec:relatedwork}

\subsection{EEG-based Affective State Estimation}

Affective state estimation from EEG recordings have gained significant interest over the past decades~\cite{al2017review}.
Majority of existing methods rely on extracting single-channel features such as statistically derived features \cite{TakahashiRemarksSignals,TangEEG-BasedSmoothing,ozdenizci2019information}, fractal dimension \cite{LiuReal-TimeEEG}, power-spectral-density (PSD) based features \cite{Lin2010EEG-BasedListening}, differential entropy \cite{ShiDifferentialEstimation} or wavelet features \cite{Akin2002ComparisonSignals}.
Multiple features across several channels are then fused to exploit the inter-channel asymmetry or connectivity relationships.
Beyond using traditional classifiers to discriminate such hand-crafted features, deep learning based end-to-end feature extraction and classification methods were also explored in EEG-based emotion recognition.
Notably, \cite{lawhern2018eegnet} introduced a deep EEG classification neural network EEGNet, as they designed a generic and compact convolutional neural network (CNN) to accurately classify EEG signals from different tasks.
Similarly specialized networks with hierarchical spatial and temporal EEG feature extraction for emotion recognition have also been proposed~\cite{li2019regional,zhong2020eeg,bethge2022domain, bethge_exploiting_2022} (see~\cite{craik2019deep} for a review).

\subsection{Synthetic EEG Data Generation}

Various approaches to generate synthetic EEG data have been recently explored to learn shared EEG components of across dataset samples. 
(cf.~\cite{ozdenizci2021use} for a recent review). One proposed framework for the generation of artificial data is generative adversarial networks (GANs)~\cite{Goodfellow:2014} which showed significant results for the generation of artificial images. This framework was applied to EEG data~\cite{Hartmann_2018,Fahimi:2019,luo2018eeg} revealing generated EEG signals by GANs resemble the temporal, spectral and spatial characteristics of real EEG.
Another line of BCI studies have used variants of variational autoencoders (VAEs)~\cite{kingma2013auto}, for unsupervised feature learning~\cite{li2019variational,aznan2019simulating,ozdenizci2021use}. 
In contrast to GANs, VAEs optimize a parameterization of a low-dimensional representation space of the training data, and hence more suitable for learning compressed data representations which is of our main interest in the affective computing applications.
Recently \cite{li2019variational} proposed to use a standard VAE to learn latent codes containing emotion-related information and use in the downstream emotion classification task via an RNN-LSTM.

\subsection{Deep EEG Latent Representation Learning}

Several approaches in EEG representation learning use deep learning models for deterministic feature learning.
One of the earlier works by \cite{bashivan2015learning} aims at finding robust representations from EEG data, that would be invariant to inter- and intra-subject differences and to inherent noise associated with EEG data collection. 
Their approach used ``EEG movies" (topology-preserving multi-spectral images) and a CNN that is applied to a cognitive load classification task.
\cite{ko2020multi} proposes a temporal and spatial feature concatenated vector representation learned with a compact deep multi-scale neural network, which is applied to diverse EEG tasks such as motor imagery, seizure or drowsiness detection.

Recently \cite{luo2020data} proposed generative VAE and GAN models for data augmentation in emotion classification. 
They show that either full or partial selection of VAE or GAN results can be used to augment EEG training datasets and demonstrate an increase in affective state classification performances. 
In contrast to our approach that utilizes raw, multi-channel EEG signals in an end-to-end manner, their method does not consider temporal dependencies in input EEG and uses hand-crafted power spectral density features as network inputs. 
Another work employs a conditional VAE model based feature encoder on EEG data, and a CNN for downstream task classification~\cite{OzdenizciTransferAutoencoders}. 
Proposed approach aims to learn subject-invariant representations by simultaneously training a cVAE and an adversarial censoring network (similar to the idea from discriminative-adversarial settings~\cite{li2021bi,ozdenizci2020learning,ozdenizci2019adversarial}), for transfer generalization of the feature encoder that can efficiently process unseen users’ EEG data for decoding. 
While prior work of EEG-based emotion recognition have mostly focused on defining explicit feature extraction and/or model architectures for detecting human emotions, our work is explicitly designed for generative-discriminative representation learning allowing for multiple classification tasks. 

\begin{figure*}[ht!]
\centering
\includegraphics[width=.99\textwidth]{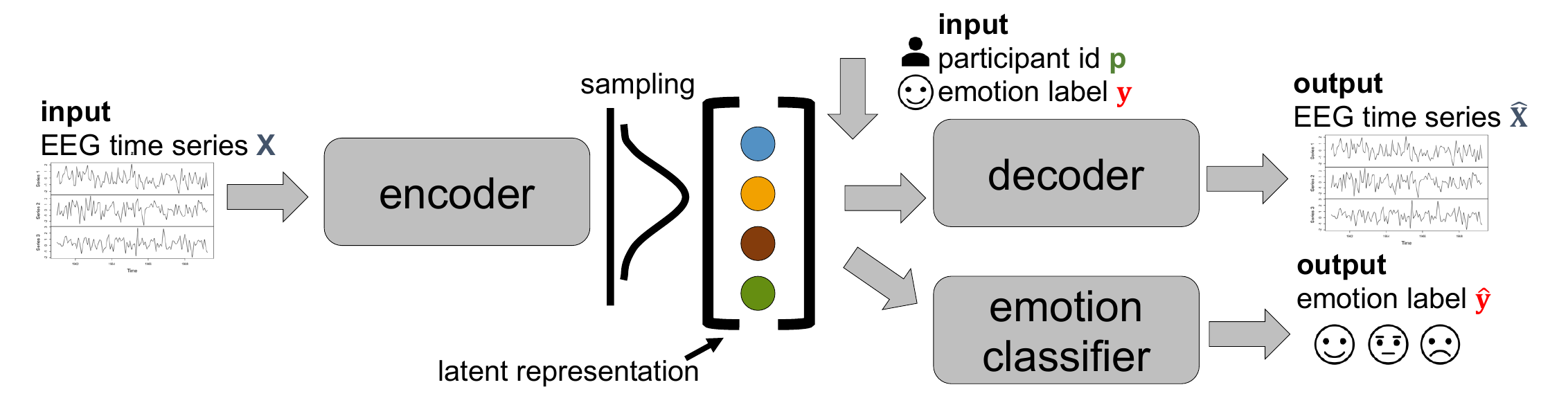}
\caption{Overview of the EEG2Vec model architecture. The variational autoencoder learns a subject- and emotion-dependent representation of EEG data that incorporates domain specific information regarding the classification task (i.e., emotion recognition). The parameters of the models are learned by minimizing the decoder reconstruction error, the loss regarding the variational posterior estimation, and the emotion classification loss.}
\label{fig:model_architecture}
\end{figure*}

\section{Methods}

\subsection{Notation}
\label{sec:notation}

We denote the labeled dataset by $\{(\bm{X}_i, y_i, p_i)\}_{i=1}^{n}$. Here $\bm{X}_i\in \mathbb{R}^{C \times T}$ defines the EEG data matrix at trial $i$ recorded from $C$ channels (i.e., EEG sensors) for $T$ discretized time samples. Accordingly, $p_i \in \{1,\ldots,P\}$ is the participant ID label and $y_i\in \{1,\ldots,L\}$ is the emotion category/class label of the corresponding trial. A deep latent representation learning model encodes an input $\bm{X}$ to learn a latent vector which we will denote by $z$.

\subsection{Conditional Variational Autoencoder (cVAE)}

In vanilla autoencoders, a deterministic encoder and decoder network pair learns a latent representation vector $z$ that is sufficient to encode the underlying shared structure across the data samples $\bm{X}_i$, such that the decoder counterpart can fully reconstruct the input samples from these learned representations.
In generative modeling with VAEs~ \cite{kingma2013auto}, however, the encoder network parameterized by $\phi$ is stochastic and estimates a true prior distribution $p(z)$ of latent $z$ via a variational posterior distribution denoted by $q_{\phi}(z|\bm{X}) \sim N (\mu_z , \sigma_z)$. In practice the encoder network estimates the parameters $\mu_z$ and $\sigma_z$, and latent vectors are obtained by sampling from the estimated variational distribution $z\sim q_{\phi}(z|\bm{X})$ at the bottleneck. The subsequent decoder network parameterized by $\theta$ is then a generative model denoted by $p_{\theta}(\bm{X}|z)$, provided with these drawn samples $z$. VAEs are trained to jointly learn better approximations of the latent prior $p(z)$ via the variational posterior $q_{\phi}(z|\bm{X})$, and successful reconstructions of $\bm{X}$ by the decoder.

In the conditional VAE (cVAE) framework \cite{sohn2015learning}, the decoder is further conditioned on at least one additional variable provided as input besides $z$. We will define a cVAE to have a decoder posterior distribution that is conditioned on both $p$ and $y$, thus modeling $p_{\theta}(\bm{X}|z,y,p)$. In a cVAE, the encoder is expected to learn representations $z$ that are invariant of $p$ and $y$, since $p$ and $y$ are already provided as input to the decoder to reconstruct the input $\bm{X}$. Training objective for variational autoencoder based models consists of maximizing a derived lower bound for the log likelihood of the data, which is usually referred as the evidence lower bound (ELBO) \cite{kingma2013auto,sohn2015learning}. Accordingly, the cVAE loss function to be minimized (i.e., negated ELBO) is given by:
\begin{equation}
\begin{split}
\mathcal{L}_{\mathbf{cVAE}} = & \underbrace{\mathbf{E}_{q_{\phi}(z|\bm{X})}[-\log  p_\theta (\bm{X}|z,p,y)]}_{\mathcal{L}_{recon}} \\ & + \underbrace{D_{\mathbf{KL}}(q_{\phi}(z|\bm{X}) || p(z) )}_{\mathcal{L}_{KL}}.
\label{eqn:cvae_loss}
\end{split}
\end{equation}
Here the first loss term $\mathcal{L}_{recon}$ corresponds to minimizing the reconstruction loss of the decoder, which is usually defined as the mean-squared-error between $\bm{X}$ and $\widehat{\bm{X}}$, and the second loss term $\mathcal{L}_{KL}$ corresponds to minimizing the Kullback–Leibler (KL) divergence between the encoder-estimated variational posterior $q_{\phi}(z|\bm{X})$ and the true distribution of $z$, which is usually defined as $p(z)\sim \mathcal{N}(0,\,I)$.

\subsection{Beta Conditional Variational Autoencoder ($\beta$-cVAE)}
\label{subsec:conditional_beta_cvae}

We extend the cVAE models with what we will refer to as a beta conditional variational autoencoder ($\beta$-cVAE). In contrast to the original $\beta$-VAEs \cite{higgins2016beta}, the model simply utilizes a conditional decoder architecture as in a cVAE. Overall, this model proposes a modification to the objective in Eq.~\eqref{eqn:cvae_loss} by introducing a hyperparameter $\beta$ for the KL-divergence term in the training loss function as follows:
\begin{equation}
\mathcal{L}_{\mathbf{\beta}\text{-}\mathbf{cVAE}} = \mathcal{L}_{recon} + \beta \mathcal{L}_{KL}.
\label{eqn:bcvae_loss}
\end{equation}

Intuitively, due to the traditional choice of $p(z)\sim \mathcal{N}(0,\,I)$, a higher $\beta$ value will converge latent representation units to more strictly follow a standard normal distribution, leading to a unit diagonal covariance matrix and (ideally) statistically independent latent representation units. Hence, during representation learning optimization, imposing a higher weight for the KL term $\beta>1$ is expected to successfully disentangle the latent representation units \cite{burgess2018understanding}. Therefore we utilize this constrained variational $\beta$-VAE \cite{higgins2016beta} approach in our conditional representation learning setting in order to impose tunable regularization of the latent space.

\subsection{EEG2Vec: A Generative-Discriminative EEG Representation Learning Framework}

In the proposed EEG2Vec framework, a conditional $\beta$-VAE and a classifier to predict $y$ (i.e., emotion category) from latent representations $z$ are simultaneously trained. We extend the objective in Eq.~\eqref{eqn:bcvae_loss} to obtain the EEG2Vec training objective function as given in Eq.~\eqref{eq:eeg2vecloss}.
\begin{equation}
\mathcal{L}_{\mathbf{EEG2Vec}} = \mathcal{L}_{\mathbf{\beta}\text{-}\mathbf{cVAE}} + \lambda \underbrace{\mathbf{E}[-\log  r_{\varphi}(y| z )]}_{\mathcal{L}_{cla}}.
\label{eq:eeg2vecloss}
\end{equation}

Here we aim to minimize the cross-entropy loss of the emotion category classification network with the additional $\mathcal{L}_{cla}$ term. The emotion predictor function is described as $r_{\varphi}$ with parameters $\varphi$, using the latent representation $z$ to predict $y$. We introduce a tunable parameter $\lambda>0$ in order to control the objective weighting for the model between a generative or discriminative behavior. For the deterministic decoder, the reconstruction loss is determined by the mean squared error of the estimated time-series EEG data. 

Given the high dimensional input EEG data $\bm{X}$, its corresponding emotion label $y$ and the participant ID $p$, the goal of the EEG2Vec model is to learn (1) a variational feature encoder $q_{\phi}(z|\bm{X})$ with parameters $\phi$ which can be generalized across subjects and emotion categories, (2) an adjacent decoder where novel data samples can be synthesized by exploiting this variational distribution, and (3) latent features from input EEG data $\bm{X}$ which are simultaneously representative in discriminating tasks or brain states associated with their corresponding emotion label $y$.

\subsection{Generative-Discriminative Inference with EEG2Vec}

After the EEG2Vec model is learned, the inference process with our model proceeds as follows. Given some participant $p$'s EEG data sample $\bm{X}$ for inference, we encode $\bm{X}$ into a latent representation distribution and obtain a sampled vector $z$. From the learned representation and the classifier network we can predict the corresponding emotion category label, and thereby performing a conventional, discriminative emotion recognition task. Furthermore we can generate synthetic EEG data samples using this sample $z$, by only providing an emotional state label $y$ and a participant ID $p$ to the decoder network due to its conditional nature. Using this manipulation scheme, we also provide a generative model that can synthesize EEG data samples specific to a particular affective state $y$ and participant ID $p$.

We regard the conditioning parameters at training and inference time to be known. However, it is possible to use the estimated values $\hat{y}$ of the emotion classifier network and the nearest neighbor $p$ of $X$ in the bottleneck as conditioning parameter estimates similarly proposed in~\cite{lopez2017conditional} as conditional values in our architecture.

\section{Experimental Study}

\subsection{Experimental Dataset}

There are a few publicly available EEG datasets with affective labels \cite{deap_database, seed_iv, katsigiannis2017dreamer, zheng2015investigating, miranda2018amigos}. 
For our study, we decided to use the STJU Emotion EEG Dataset (SEED) dataset \cite{zheng2015investigating} as it uses a rather simple labeling system with three distinct classes: negative, neutral and positive. 
This facilitates the learning of embeddings and reduces complexity for the emotion classifier. 
The dataset contains 62-channel EEG recordings sampled at 1000Hz from 15 participants recorded from 3 sessions. 
During each session, the participants were shown 15 film clips that should elicit either negative, neutral, or positive emotions representing the emotional label for that trial. 

Since the duration of each experiment was different, to unify, we determined a 185 seconds duration (being the shortest duration of all experiments) as the standard experiment duration. For those experiments which duration is longer than 185s, the last 185s segment were selected. In order to avoid the possible interference or the possible emotions has not been elicited at the beginning of the experiment, we removed the first 30 seconds of EEG data, (i.e., only 155 seconds of segments were used~\cite{qing2019interpretable}). Data were further preprocessed as in~\cite{zheng2015investigating,qing2019interpretable} and accordingly first downsampled to 200Hz, and then a 2-40 Hz Butterworth bandpass filter was applied for low and high frequency band artifacts. EEG data were segmented into 2 second non-overlapping time intervals in accordance with previous work~\cite{candra2015investigation}. Finally, data is normalized to the range of $[0,1]$. No offline channel selection was performed.

\subsection{EEG2Vec Model Specifications}

We developed our feature encoder backbone based on the well-known convolutional EEGNet architecture \cite{lawhern2018eegnet} due to its multi-purpose EEG representation learning capabilities. We modified the architecture based on the input representations of our dataset (e.g., sampling rate or number of EEG sensors). The decoder is implemented using inverse versions of the encoder layers. The encoded latent representation $z$ is used as an input to the classifier that aims to accurately predict the corresponding emotional state. Herein, $z$ is propagated through three fully-connected layers where the last layer is equipped with a softmax activation function. 
We initially set $\beta=\lambda=1$, such that the reconstruction and classification performances are equally weighted in the loss function. Depending on the practical applications of our embedding, we can set these parameters accordingly.

\subsection{Model Training and Evaluation}
\label{subsec:model_training_and_evaluation}

\subsubsection{Implementation}
\label{subsec:exp_study_implementations}

Networks were trained with 50 training trials per batch for at most 2000 epochs with early stopping based on the model loss on the validation set. Parameter updates were performed once per batch with Adam. The input EEG data matrices are of dimensions $C=62$ times 400 discretized time samples. Dimensionality of latent $z$ was determined as 1000.
We formulate $p$ as a one-hot encoded vector (i.e., a $P$-dimensional vector with a value of 1 at the s'th index and zero in other indices) and $y$ to be the one-hot encoded emotion class vector. We use both $p$ and $y$ as conditioning parameters for the decoder to enforce the learning of subject- and emotion-dependent generated EEG.
We used the TensorFlow libraries with the Keras API. 

\subsubsection{Neural Network Architectures}

To measure the effectiveness of our results we measure both signal reconstruction ability and emotion classification performance against a baseline discriminative model.
We report the emotion classification results of a discriminative EEGNet~\cite{lawhern2018eegnet} model, which we used as the backbone in an accuracy trade-off to also realize generative EEG data modeling.

\subsubsection{Evaluation}
\label{subsubsec:evaluation}

Our experiments evaluate the performance of EEG2Vec in comparison to a state-of-the-art discriminative baseline model in terms of emotion state classification, and also demonstrates its EEG signal reconstruction ability. We initially separate a holdout fixed testing set consisting of $10\%$ of the complete dataset. Then we split the remaining $90\%$ of the data into training and validation sets by 5-fold cross-validation. We ensure that in both training, validation and testing sets, all affective states (i.e., classes) and subjects are represented equally in terms of number of data samples. 

\section{Experimental Results}

\subsection{Learning Deep Latent Representations}

\begin{figure*}
    \begin{minipage}[b]{.48\textwidth}
    \includegraphics[width=.99\linewidth]{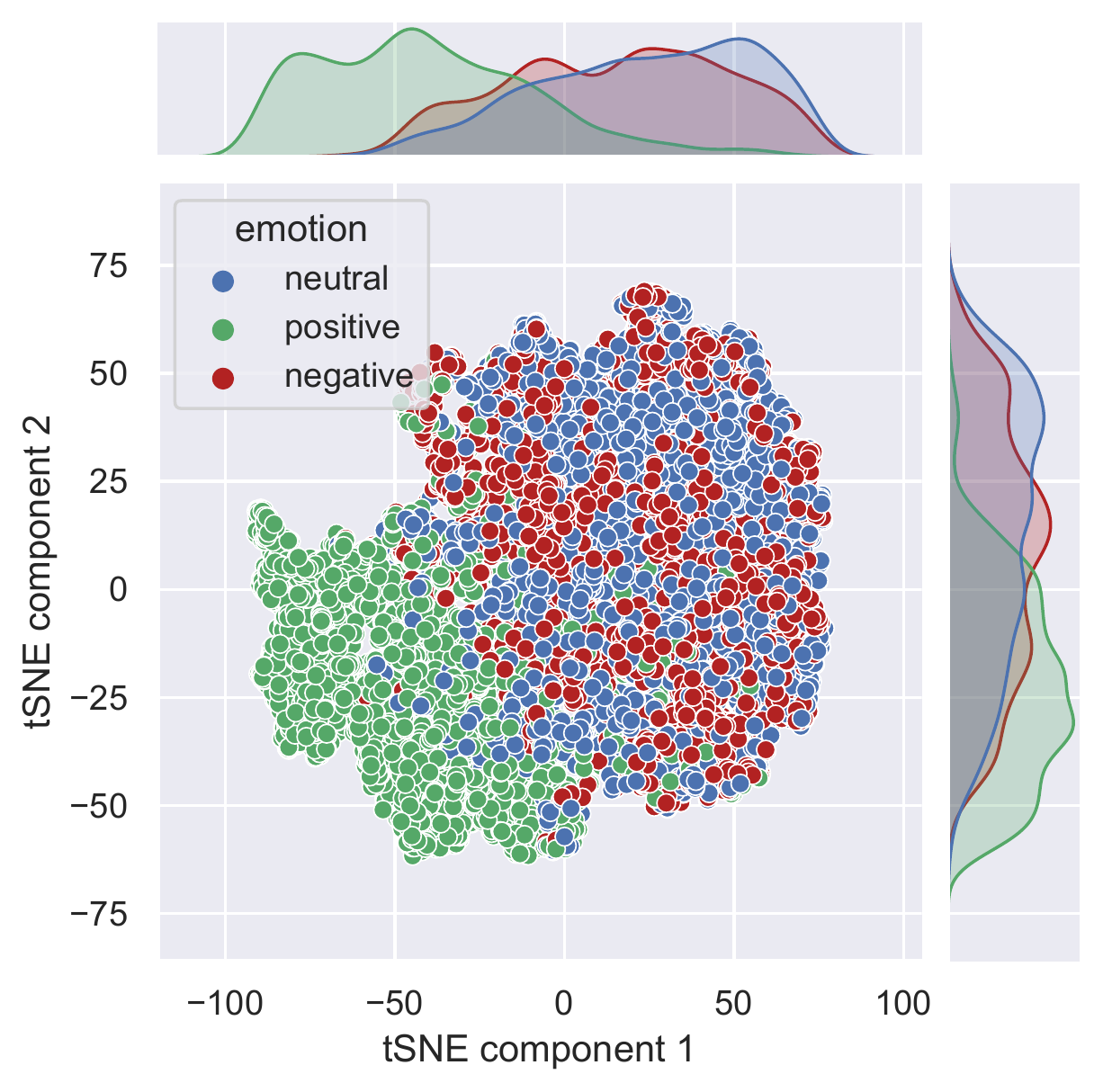}
    \caption{Latent space visualization of learned representation $z$ using a t-distributed Stochastic Neighbor Embedding (t-SNE) with two components. We used the encoder network of the EEG2Vec model ($\beta=1$) to transform all observations from the validation set to $z$.}
        \label{fig:z_tsne_beta=1}
    \end{minipage}
    \qquad
    \begin{minipage}[b]{.48\textwidth}
    \includegraphics[width=.99\linewidth]{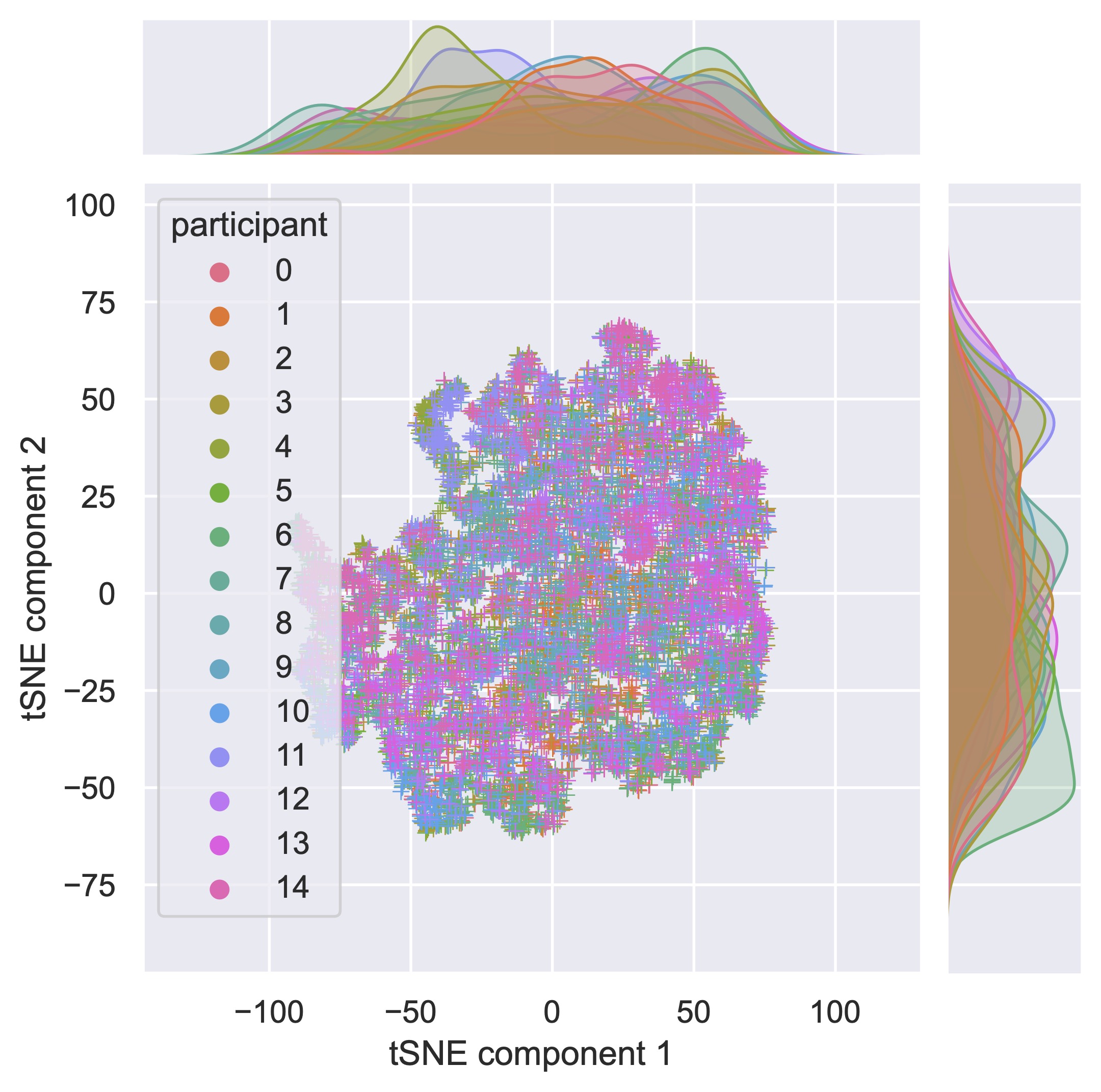}
    \caption{Latent space visualization of learned representation $z$ using a t-distributed Stochastic Neighbor Embedding (t-SNE) with 2 components. Colors represent different participant IDs.}
        \label{fig:z_tsne_beta=1_participant}
    \end{minipage}
\end{figure*}

We first investigate the structuring of the learned embedding. Figure~\ref{fig:z_tsne_beta=1} visualizes the 1000-dimensional learned embedding $z$ into a two-dimensional scatterplot via t-distributed Stochastic Neighbor Embedding (t-SNE) \cite{maaten2008visualizing}. All tSNE visualizations are generated with default parameters: perplexity $=30$, number of iterations $= 1000$.
We observe a discriminative pattern of different emotion types in $z$ as the EEG data of positive emotions is found prevalently in the left half of the scatterplot. This indicates that the encoder can embed affective state information in the latent representation, and the auxiliary classifier can predict emotions easier from this representation (see Section~\ref{sec:exp_results_emotion_classification}). Thus $z$ incorporates specific information about the emotion category. However, we also observe that data from negative and neutral emotion class are overlapping in Figure \ref{fig:z_tsne_beta=1}, which is further investigated in Section~\ref{sec:exp_results_emotion_classification} and mainly due to harder discriminability \cite{liu2016multimodal} and changing spatio-temporal patterns for these emotions affecting the input EEG.

%

Figure \ref{fig:z_tsne_beta=1_participant} demonstrates similar t-SNE embeddings of $z$ separated by the participants. We observe no particular global clustering in $z$ based on the participant. This is an expected behavior as the participant ID is provided to the decoder network as an additional input besides $z$ and hence $z$ is expected to be invariant to participant IDs.


Participant-dependent affective state information is encoded into a low-dimensional space with 1000 dimensional latent variables, which is only $4.03\%$ of the original EEG data size ($62 \times 400$). This compressed representation can thereby be efficiently used for EEG signal processing with low computational cost and memory requirements as demonstrated practically with similar techniques in~\cite{li2013continuous,shin2015simple}.

\subsection{Emotion Classification from EEG}
\label{sec:exp_results_emotion_classification}

\begin{figure}[t]
    \centering
    \includegraphics[width=0.99\linewidth]{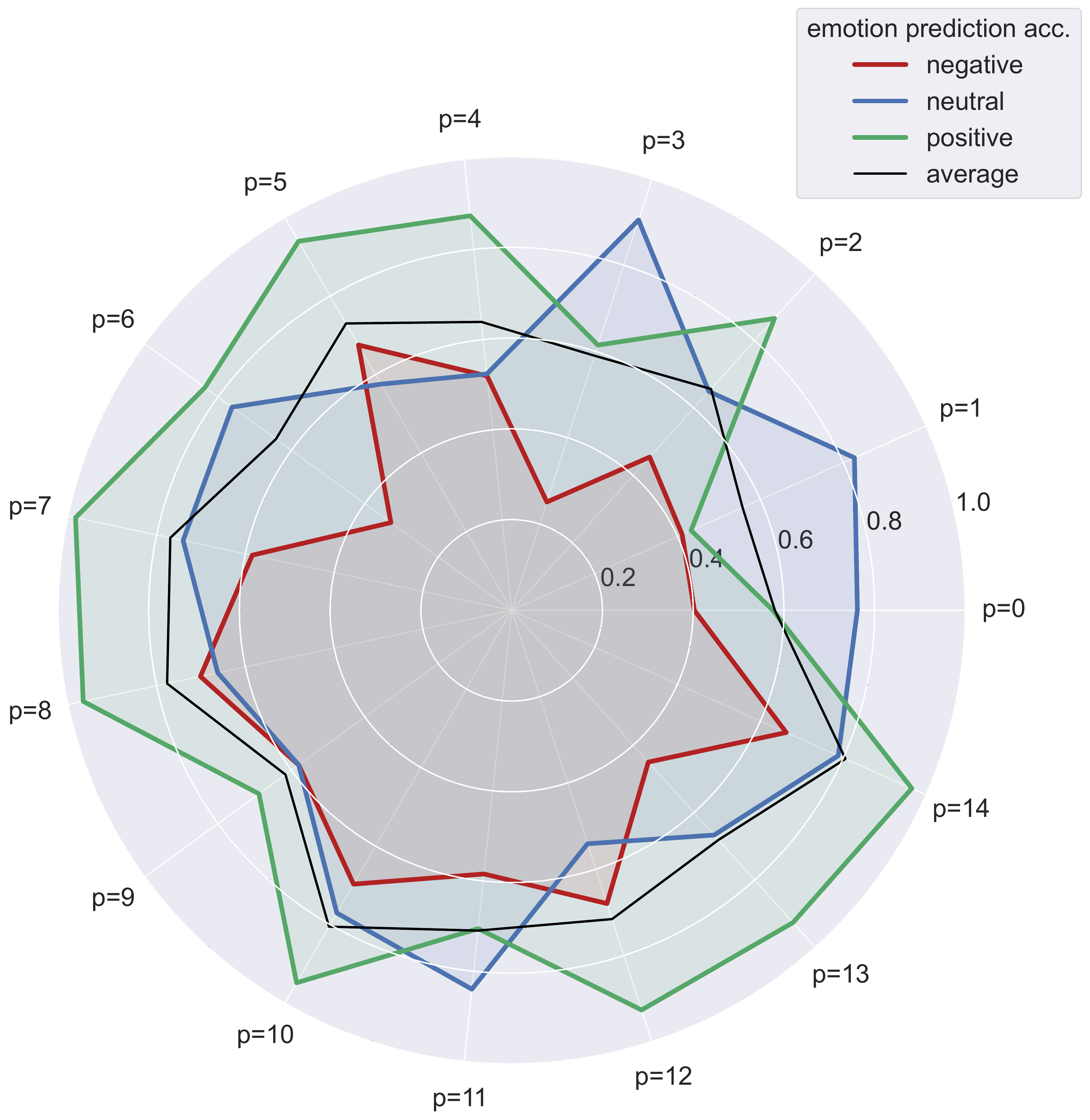}
    \caption{Emotion prediction accuracy of EEG2Vec for different participants in the testing set. Different colored polar-plots represent the achieved accuracy of the prediction of the specific emotion class. Black curve indicates the mean accuracy per participant over all emotional classes.}
    \label{fig:participant_wise_emotion_prediction}
\end{figure}

Within our multi-outcome EEG2Vec framework (both reconstructing EEG and predicting emotional labels), we weigh the importance of signal reconstruction and emotion classification equally via setting $\lambda$ appropriately. We train our model as described in Section~\ref{subsec:model_training_and_evaluation}. We demonstrate the differences in emotion classification across individual participants in Figure~\ref{fig:participant_wise_emotion_prediction}.

Our model is able to achieve a $68.49\%$ testing classification accuracy (significantly above three-class decoding chance-level accuracy with $p=10^{-4}$ using a Wilcoxon signed rank test), and reliably predict positive emotions with a very low false-prediction error rate: very high precision of $82\%$ and recall of $89\%$. Figure \ref{fig:participant_wise_emotion_prediction} further supports the existence of a significantly larger prediction performance for positive affective state classification (i.e., green curve) on an individual participant-level as well. These results are computed on the basis of a latent representation invariant of participant information and thereby represent a participant-independent emotion recognition performance.

On the other hand, the fully-discriminative EEGNet baseline model is able to achieve an affective state prediction accuracy of $77.27\%$ on the same testing set. The increase in classification accuracy with EEGNet can be simply explained by using a sole model objective of deterministic affective state classification, versus learning robust variational embeddings for emotion prediction and EEG data augmentation simultaneously as in our EEG2Vec model. Compared to the feature-optimized emotion detectors such as \cite{zheng2015investigating} or more general deep learning approaches such as EEGNet, EEG2Vec compensates a slightly lower classification accuracy while being able to provide a low-dimensional, affect-distinctive representation. However our model classification performance converges to the overall EEGNet accuracy as $\lambda$ increases, i.e., weighing a stronger classification performance more than generative capabilities of the model.

\subsection{Condition-Specific Artificial EEG Synthesis}

\begin{figure}[t!]
\centering
\includegraphics[width=1\linewidth]{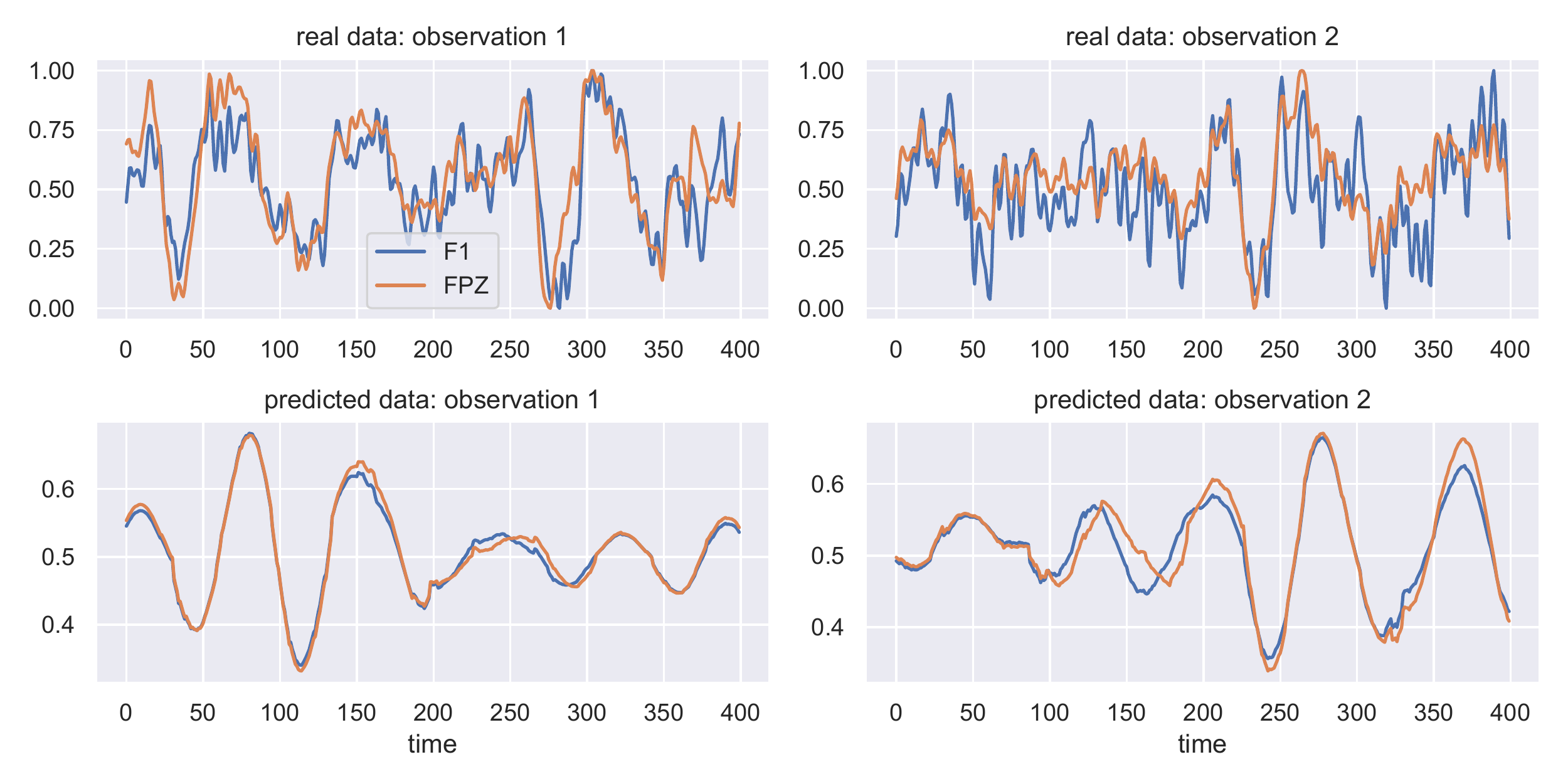}
\caption{Real sample versus generated time-series EEG data for two EEG channels F1 and FPZ. Observation 1 has positive affective state, whereas observation 2 is of negative affective state, both from the same participant.}
\label{fig:gen_data}
\end{figure}

Figure~\ref{fig:gen_data} visualizes generated EEG examples, where sampled latent representations $z$ are exploited with the decoder network for reconstruction.
To further validate the usefulness of the synthetic EEG data, we employ an evaluation of an emotion classification on original and synthetic data in Table \ref{tab:syntheticdataevaluationforemotionprediction}. In the analysis of the prediction performance, we observe that original data merged with $20\%$ synthetic data from our model can improve classification accuracy by several percentage points (from $66\%$ to $69\%$) vs. only using the original data. We detect an overall modest increase in many participants' classification accuracy with the synthetically boosted model ($20\%$ synthetic data), however we see the largest increase in emotion recognition increased by $42.85\%$ by participant 12's and $31.57\%$ when looking at participant 11, while the largest decrease in accuracy of $-17.02\%$ is observed when decoding the data of participant 4.

High increases in performances for specific subjects using EEG2Vec's synthetic data augmentation technique are likely due to the subject- and session-variant nature of EEG data which is learned by our model. This phenomenon is similar to the work by~\cite{aznan2019simulating}, which showed that generating supplementary synthetic EEG improved steady state visually evoked potential classification across subjects. We opted against using a higher percentage of synthetic data since the models can then easily overfit to the distribution of the synthetic data. Overall, our results show, that EEG2Vec's synthetic EEG data is beneficial to overall increase the detection ability of newly trained models and can boost the emotion recognition performance of particular subjects. 

\begin{figure}[t!]
\centering
\includegraphics[width=.99\linewidth]{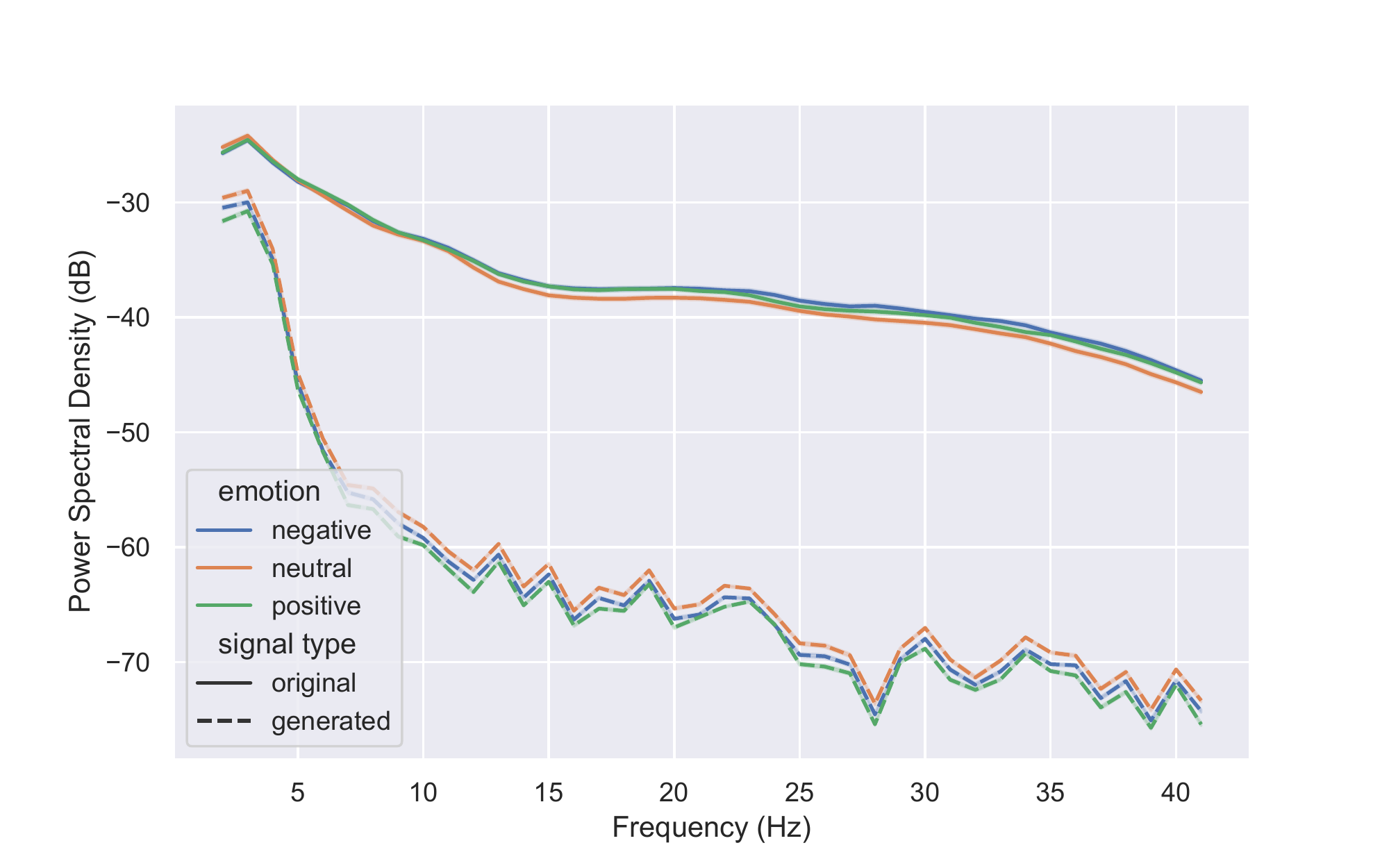}
\caption{Mean PSD of original and generated EEG data (channel: F3, validation set) with EEG2Vec. Affective states are depicted in different colors. Low frequency can be generated accurately, while higher frequency signals cannot be appropriately synthesized as can be seen from the diverging lines of reconstructed and original PSD-Frequency lines.}
\label{fig:psd_mean_channel3}
\end{figure}

Our architecture is able to generate visually-representative EEG data with relevant low-frequency components.
Looking at the power-spectral density (PSD) plot in Figure~\ref{fig:psd_mean_channel3}, the reconstruction ability for lower frequency ($\le10Hz$) of the original EEG signal is good, because at a low-frequency of $~1Hz$ both reconstructed and original signal starting with a PSD of -25 to -30 dB followed by the same downward PSD curve trend. 
Here, the mean PSD is calculated by averaging over all observation EEG sequences the Welch's power-spectral density results with frequency min/max of 2/41 Hz, FFT length of 200 and 50 observations overlap.
However, higher frequency ($>10Hz$) cannot be learned appropriately as the PSD values differ heavily (generated PSD values range from -60 to -75 dB whereas the original signal are from -35 to -40 dB). 
This can be partly explained by the up-sampling components of the generator network which introduce aliasing frequency artifacts \cite{Hartmann_2018} as well as noise artifacts in the inputs are not captured.


\begin{table}[t!]
\centering
\caption{Evaluation of synthetic EEG2Vec data that is additionally used for classification of emotional states.}
\label{tab:syntheticdataevaluationforemotionprediction}
\resizebox{\linewidth}{!}{%
\begin{tabular}{l|cccc}
\toprule
\multicolumn{1}{c|}{\textbf{model} }   & \textbf{accuracy}             & $\boldsymbol{F_1}$                & \textbf{precision}             & \textbf{recall}                 \\
                               & \multicolumn{1}{l}{} & \multicolumn{1}{l}{} & \multicolumn{2}{c}{negative / neutral / positive} \\ 
                               \midrule
original                       & .66                  & .66                  & .55/.60/.83             & .52/.63/.82             \\
original + 5\% synthetic data  & .68                  & .68                  & .53/.56/.82             & .65/.56/.82             \\
original + 20\% synthetic data & .69                  & .65                  & .51/.57/.71             & .68/.57/.71            \\
\bottomrule
\end{tabular}%
}
\end{table}



\subsection{Effects of Disentangling in Latent Space}

Lastly, we examine reconstruction performance of EEG2Vec by varying the KL-divergence factor $\beta$ while regularizing the latent space. 
With increasing $\beta$, the factors of the learned embedding are directed towards a more disentangled form, i.e., statistical independence than on reconstruction, which results in more uncorrelated latent features with less reconstruction ability.
If each variable in the inferred latent representation $z$ is only sensitive to one single generative factor and relatively invariant to other factors, we will say this representation is disentangled. 
Having disentangled representations can help to reduce information overlay from different factors, and promote better interpretability and easier generalization to a variety of tasks \cite{higgins2016beta}.

\begin{figure}[t!]
    \centering
    \includegraphics[width=0.94\linewidth]{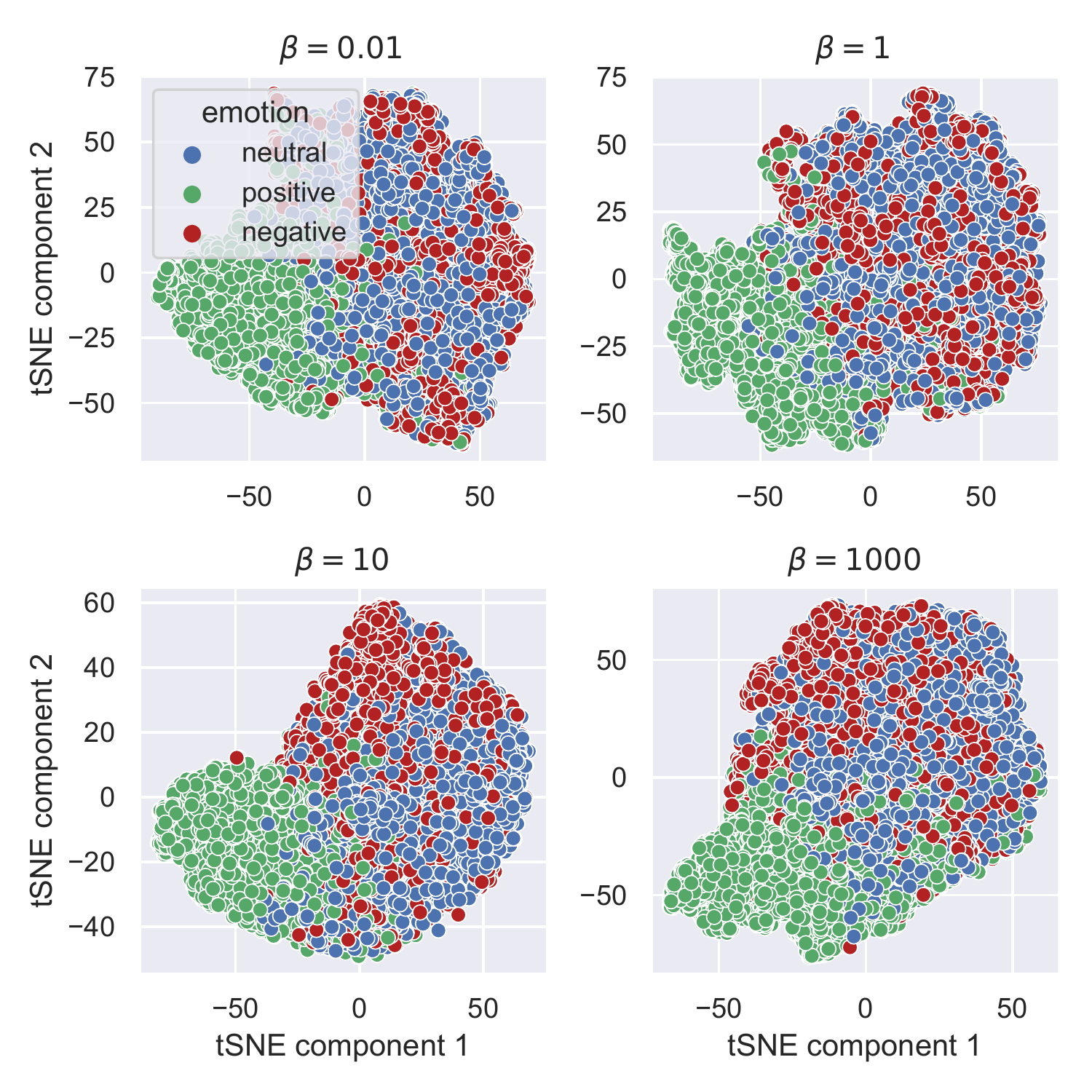}
    \caption{t-SNE visualizations of latent embeddings $z$ for varying latent space regularization weights $\beta$. }
    \label{fig:beta_variation_z_emotion_window_size=400}
\end{figure}

For all $\beta$ values depicted in Figure~\ref{fig:beta_variation_z_emotion_window_size=400}, $z$ is showing positive emotions in a discriminative region. 
We also observe that with higher $\beta$ (lower plots) negative and neutral affective state $z$ are showing more discriminative clusters which are depicted more clearly in different subspaces.
Results indicate that higher disentanglement can favor higher distinction between affective states of the clusters.

\section{Discussion}

We propose EEG2Vec as a mechanism to learn latent representations of affective EEG data that allow for general use in various generative and discriminative machine learning paradigms.
Our model learns vectorized representations (i.e., \textit{embeddings}) of EEG responses to emotional videos that are discriminative of the affective states, as well as sufficiently representative to generate synthetic EEG data. 
In doing so, learned \textit{embeddings} can also be used to generate synthetic EEG data that is both participant- and emotion-specific, simply by sampling from the latent state probability function. 
Our results altogether show that the proposed architecture is able to learn both efficient (lower dimensionality with $z$) and expressive (able to maintain useful properties with $z$) representations.
One important limitation of our approach lies on the accessible training dataset infrastructure. It is naturally likely that the amount of participant-specific data can impact optimization if not accounted for.
So far we only considered learning from a balanced training data set in terms of participant IDs and class labels by stratifying our available training set size.
Nevertheless the proposed EEG2Vec pipeline with sufficient amount of data allows future research to exploit low-rank EEG representations with less memory demand for general purpose edge applications (e.g., wearable computing \cite{han2020disentangled,han2021universal} or human-robot interaction~\cite{Ozdenizci:2018EMBC}).

\section*{Acknowledgments}
\addcontentsline{toc}{section}{Acknowledgments}
This research was partly funded by the Deutsche Forschungsgemeinschaft (DFG, German Research Foundation) in TRR161 (Quantitative methods for visual computing, Project ID 251654672) in Project C06. LLC is supported by the research initiative “Instant Teaming between Humans and Production Systems" co-financed by tax funds of the Saxony State Ministry of Science and Art (SMWK3-7304/35/3-2021/4819).

\bibliographystyle{IEEEtran}
\bibliography{bibliography}

\end{document}